# Automatic segmentation method of pelvic floor levator hiatus in ultrasound using a self-normalising neural network

Ester Bonmati,[a,b,c,*] Yipeng Hu,[a,b,c] Nikhil Sindhwani,[d] Hans Peter Dietz,[e] Jan D'hooge,[d] Dean Barratt,[a,b,c] Jan Deprest,[b,d] Tom Vercauteren[a,b,c,d]

[a] UCL Centre for Medical Image Computing, University College London, London, UK
[b] Wellcome/EPSRC Centre for Interventional and Surgical Sciences, University College London, London, UK
[c] Department of Medical Physics and Biomedical Engineering, University College London, London, UK
[d] Department of Development and Regeneration, Cluster Urogenital Surgery and Clinical Department of Obstetrics and Gynaecology, University Hospitals Leuven, KU Leuven, Leuven, Belgium
[e] Sydney Medical School Nepean, Nepean Hospital, Penrith, Australia

**Abstract**. Segmentation of the levator hiatus in ultrasound allows to extract biometrics which are of importance for pelvic floor disorder assessment. In this work, we present a fully automatic method using a convolutional neural network (CNN) to outline the levator hiatus in a 2D image extracted from a 3D ultrasound volume. In particular, our method uses a recently developed scaled exponential linear unit (SELU) as a nonlinear self-normalising activation function, which for the first time has been applied in medical imaging with CNN. SELU has important advantages such as being parameter-free and mini-batch independent, which may help to overcome memory constraints during training. A dataset with 91 images from 35 patients during Valsalva, contraction and rest, all labelled by three operators, is used for training and evaluation in a leave-one-patient-out cross-validation. Results show a median Dice similarity coefficient of 0.90 with an interquartile range of 0.08, with equivalent performance to the three operators (with a Williams' index of 1.03), and outperforming a U-Net architecture without the need for batch normalisation. We conclude that the proposed fully automatic method achieved equivalent accuracy in segmenting the pelvic floor levator hiatus compared to a previous semi-automatic approach.

**Keywords**: levator hiatus, automatic segmentation, SNN, ultrasound, CNN

*Ester Bonmati, E-mail: e.bonmati@ucl.ac.uk

## 1 Introduction

Pelvic Organ Prolapse (POP) is the abnormal downward descent of pelvic organs including, i.e., the bladder, uterus and/or the rectum or small bowel, through the genital hiatus, resulting in a protrusion through the vagina. In a previous study, 27,342 women between the age of 50-79 years, were examined and found that about 41% showed some degree of prolapse[1]. Ultrasound is at present the most widely used imaging modality to assess the anatomical integrity and function of pelvic floor, because of availability and non-invasiveness. Since the levator hiatus is the portal through which POP must occur, its dimensions and appearance are measured and recorded during an ultrasound exam. The hiatal dimensions have also been correlated with severity of prolapse, levator muscle avulsion and even prolapse recurrence after surgery[2–4].

During a transperineal ultrasound examination, 3D volumes are acquired during Valsalva manoeuvre (act of expiration while closing the airways after a full inspiration), at pelvic floor muscle contraction and during rest. The hiatal dimensions and its area are then recorded by manually outlining the levator hiatus in the oblique axial 2D plane at the level of minimal anterioposterior hiatal dimensions (referred to as the C-plane hereinafter)[2].

The main limitation of this technique is the high variability between operators in assessing the images and the operator time required. Sindhwani et al.[5] earlier proposed a semi-automatic method to segment the levator hiatus in a predefined C-plane. In order to define the C-plane, their approach requires first the identification of two 3D anatomical landmarks within the 3D volume, the posterior aspect of the symphysis pubis (SP) and the anterior border of the pubovisceral muscle (PM), which are labelled manually. Then, the SP and PM are manually defined on the selected C-plane and the system performs the outlining automatically. Although it is true that most of the times the SP and PM defined in the 3D volume may correspond in the 2D image, this is not always the case and may need to be corrected in the axial view. Therefore, Sindhwani et al.[5] method, requires to identify the two points in both images. Additionally, the contours in the C-plane rely on the manual addition of a third point and may require some additional manual adjustments. This method was shown to reduce interoperator variability in comparison to manual segmentation. Overall, despite interesting results, the procedure still lacks of automation, limiting its reproducibility, and requires operator inputs and, consequently, time.

Recently, convolutional neural networks (CNN) have been shown to be able to successfully perform several tasks such as classify, detect or segment objects in the context of medical image analysis[6]. Litjens et al.[7] provide a good review on deep learning in medical image analysis. To segment medical images, different deep learning approaches have been proposed in 2D (e.g. left and right ventricle[8], liver[9]), in 3D (e.g brain tumour[10], liver[11]) and have recently been extended to support interactive segmentation in both 2D and 3D[12,13]. In particular, using 2D ultrasound images, CNN have been employed to successfully segment deep brain regions[14], the foetal abdomen[15], thyroid nodule[16], foetal left ventricle[17], and vessels[18] providing a fully automatic approach.

In this work, we propose a fully automatic method to segment, in manually defined 2D C-planes, the levator hiatus from ultrasound volumes thereby further automating the process of outlining the pelvic floor. In particular, we employ a self-normalising neural network (SNN) using a recently developed scaled exponential linear unit (SELU) as a nonlinear activation function, with and without SELU-dropout[19], showing competitive results compared to the equivalent network not using SELU. To the best of our knowledge, our work is the first attempt to combine SELU with CNN. SNNs have clear benefits in many medical imaging applications. These include the parameter-free and mini-batch independence nature of SNNs. In deep learning for medical imaging applications, memory constraints are frequently reached during training. Having opportunities to reduce the complexity of the network and being able to use a smaller mini-batch size (in contrast to batch normalisation), without sacrificing the generalisation performance, are both crucial for many applications.

We train and evaluate the network using 91 C-plane ultrasound images, from 35 patients, in a leave-one-patient-out cross-validation. The dataset contains images at three different stages: full Valsalva, contraction and rest. For each image, three labels from three different operators are available and are used for training and evaluation within the cross-validation experiment. Furthermore, we directly compare the results using U-Net-based architectures[20,21], a ResNet approach[22], and the proposed network with and without SELU-dropout.

## 2    Method

*2.1 Self-normalising neural networks for ultrasound segmentation*

In this work, segmenting anatomical regions of interest in medical images is posed as a joint classification problem for all image pixels using a convolutional neural network. Ultrasound images, which contain relatively sparse features that are depth- and orientation-dependent representation of the anatomy, pose a challenging task for traditional CNNs. Therefore, the appropriate regularisation and robustness of the training may be important to successfully segment ultrasound images. In recent years, rectified linear units, has become the *de facto* standard nonlinear activation function for many CNN architectures due to its simplicity and provides partially constant, non-saturating gradient, while batch normalisation, retains a similar importance by effectively reducing the internal variate shift and therefore regularises and accelerates the network training[23]. However, the stochastic gradient descent with relatively small data and mini-batch sizes (commonly found in medical image analysis applications) may significantly perturb the training so that the variance of the training error becomes large. This has also been reported by the training error curves from previous work[24]. This work explores an alternative construction of the nonlinear activation function used in a self-normalising neural network, a recent development suggesting to use a scaled exponential linear unit (SELU) function[19]. The proposed SELU constructs a particular form of parameter-free scaled exponential linear unit so that the mapped variance can be effectively normalised, i.e. by dampening the larger variances and accelerate the smaller ones. As a result, batch-dependent normalisation may not be needed, which means that there is no mini-batch size limitation and networks should be able to obtain equivalent results with reduced memory constraints. The SELU activation function is defined as:

$$SELU(x) = \lambda \begin{cases} x & if\ x > 0 \\ \alpha e^x - \alpha & if\ x \leq 0 \end{cases}, \qquad (1)$$

where scale $\lambda = 1.0507$ and $\alpha = 1.6733$ (see Klambauer et al.[19] for details on the derivation of these two parameters). This specific form in Equation 1 ensures the mapped variance by the SELU activation is effectively bounded [19] thereby leading to a self-normalising property.

*2.2 Network architecture*

We adapt a U-Net architecture[20,25] as a baseline CNN to assess the segmentation algorithms. We refer to the proposed self-normalising U-Net-based network as SU-Net hereinafter. The detailed network architecture is illustrated in Fig. 1. Each block consists of two convolutions, with a kernel size of 2x2, each followed by a SELU activation. Down-sampling is achieved with a max-pooling with a kernel size of 2x2 and stride 2x2 which halves the sizes of the feature maps preserving the number of channels, while up-sampling doubles the feature map sizes, also preserving the number of channels. Up-sampling is performed by a transposed convolution with a 2x2 stride. After each up-sampling, the feature maps are concatenated with the last feature maps of the same size (before pooling). The last block contains an extra convolution and the corresponding SELU activation. As shown in Fig. 2, all the batch normalisation with rectified linear units (ReLU) blocks are replaced by a single SELU activation (described in Section 2.1). For the case of SU-Net with SELU-dropout, the dropout was applied after each convolution. SELU-dropout works with scaled exponential linear units by randomly setting activations to the negative saturation value (in contrast to zero variance in ReLU), in order to keep the mean and variance. The weighted sum of a L2 regularisation loss with of the probabilistic Dice score using label smoothing is used as a loss function[26,27].

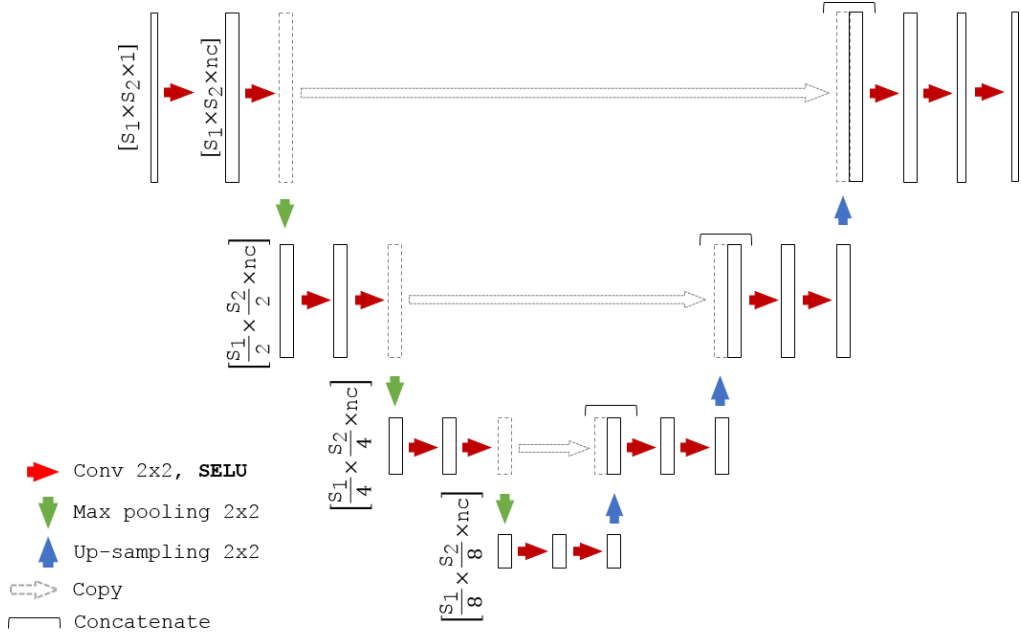

**Fig. 1** Network architecture, where $S_1$ and $S_2$ correspond to the spatial dimension and *nc* to the number of channels. For the U-Net, the SELU unit is replaced by batch normalisation and ReLU, and for the U-Net with dilated convolution (U-Net+DC), the last layer is also replaced by a dilated convolution.

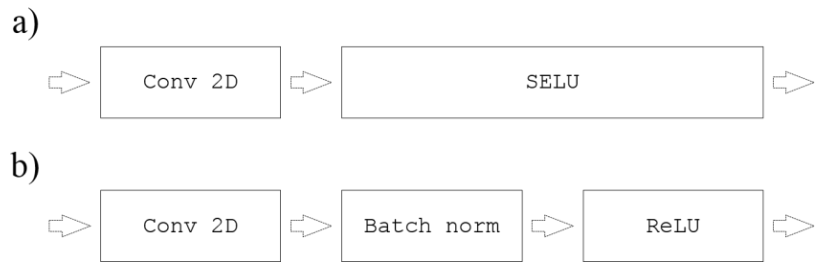

**Fig. 2** SU-Net architecture (a) versus a U-Net architecture (b).

*2.3 Networks evaluation*

Manually labelled ultrasound images, each of which labelled by three individual operators, are available to train the networks. Our benchmark include the proposed SU-Net using SELU (SU-Net), the SU-Net also using SELU-dropout (SU-Net+dropout), and a baseline U-Net using batch normalisation and ReLU (U-Net) sharing the same architecture as the SU-Net (**Fig.** 1). Other hyper-parameters are kept fixed for all these architectures. Additionally, similar to Vigneault et al.[25], we also compare the results with a U-Net in which the last layer convolutions are replaced by dilated convolutions (U-Net+DC), and with a ResNet architecture[22]. Hyper-parameters used in the implementation of the U-Net+DC and ResNet networks are described in Section 3.2 Evaluation is performed in a leave-one-patient-out cross-validation, in which the networks are trained 35 times using data from 34 patients while the contours from the different images of the left-out patient are used in testing. As a result, 91 automatic segmentations are obtained from the 35-fold validation, corresponding to the size of the original dataset.

*2.4 Metrics*

Results are evaluated using two region-based measures: Dice similarity coefficient[28] and Jaccard coefficient[29], and two distance-based measures: symmetric Hausdorff distance and mean absolute distance (*MAD*). The choice of this comprehensive set of metrics aims to allow direct comparison with the results from a previous study using the same dataset[5]. Additionally, we include two more region-based measures, the false positive Dice (*FPD*) and the false negative Dice (*FND*)[30] and one distance-based measure, the symmetric mean absolute distance (*SMAD*) which is the symmetric version of *MAD*.

Let *A* and *B* be two binary images which correspond to two labelled levator hiatus, in our evaluation, *A* corresponds to an automatic segmentation and *B* to a manual segmentation (ground truth), the Dice similarity coefficient $D(A,B) = 2|A \cap B|/(|A| + |B|)$ expresses the overlap or similarity between label $A$ and $B$. The Jaccard coefficient $J(A,B) = |A \cap B|/|A \cup B|$ provides an alternative, more conservative overlap measure between $A$ and $B$. $FPD = 2|A \cap \bar{B}|/(|A| + |B|)$ and $FND = 2|\bar{A} \cap B|/(|A| + |B|)$ where $\bar{A}$ refers to the complement of $A$ and $\bar{B}$ to the complement of $B$, and can be used to quantify if the method is over-segmenting or under-segmenting, respectively.

Let $X = \{x_1, x_2, \ldots, x_n\}$ and $Y = \{y_1, y_2, \ldots, y_n\}$ be two finite 2D point sets sufficiently sampled from the contours or boundaries of binary images $A$ and $B$ with sizes $n_x$ and $n_y$, respectively, the symmetric Hausdorff distance ($H$) finds the maximum distance between each point of a set to the closest point of the other set as follows: $H(X,Y) = \max\{\max\{|d(x,Y)|\}, \max|d(y,X)|\}, \forall x \in X, \forall y \in Y$, where $d(x,Y) = \min\{\|x - y_i\|\}, i = \{1..n_y\}$ and $\|x - y_i\|$ is the Euclidean distance between the 2D point $x$ and the $i$th point of $Y$. This measure quantifies the maximum level of disagreement between two labels. The mean absolute distance $MAD(X,Y) = \sum_{i=1}^{n_x}|d(x_i,Y)|/n_x$, quantifies the averaged level of agreement between contours $X$ and $Y$ by finding the averaged distance between all points of a set to the closest point of the other set. Note that, as previously mentioned, $MAD$ is asymmetric, therefore we also include the symmetric mean absolute distance $SMAD(X,Y) = \frac{1}{n_x+n_y}\left(\sum_{i=1}^{n_x}|d(x_i,Y| + \sum_{i=1}^{n_y}|d(y_i,X)|\right)$.

*2.5 Statistical comparative analysis*

Performance is quantified and compared by evaluating the computer-to-observer differences (i.e., the agreement between the automatic segmentation and the manual segmentations). A pairwise comparison approach between each label obtained with the automatic method and the three labels available for each image is performed by considering all the metrics described in Section 2.4

Performance quantification is presented for all network architectures described. Furthermore, statistical analysis employing a paired two-sample Student's t-test is used to test whether the differences in performance between SU-Net and U-Net, U-Net+DC, ResNet and SU-Net+dropout are statistically significant different.

Using a similar pairwise approach, interobserver differences (i.e., agreement between manual segmentations from the three operators) are quantified to allow a further comparison with the automatic methods.

The extended Williams' index is a statistical test for numeric multivariate data to test the null hypothesis that the automatic method agrees with the three operators, and that the three operators agree with each other[31,32]. This index quantifies the ratio of agreement by calculating the number of times that the automatic boundaries are within the observer boundaries. If the 95% confidence interval (CI) of the Williams' index contains the value 1.0, it implies that the test fails in rejecting the null hypothesis that the agreement between the automatic method and the three operators is not significantly different. We test the level of agreement between the automatic and manual segmentations based on the metrics defined in Section 2.4

*2.6 Clinical impact*

The dimensions of the levator hiatus on ultrasound is a biometric measurement used to assess the status of the levator hiatus, and is associated both with symptoms and signs of prolapse as well as with recurrence after surgical treatment[2]. Therefore, we extend the analysis to include the area measurement from the manual and automatic segmentations, in order to provide further clinical relevance in assessing the segmentation algorithms. Evaluation is performed by grouping the images in the three different stages: during rest, Valsalva and contraction. Williams' index is again used to test the level of agreement between the automatic and manual labels.

# 3 Experiments

## 3.1 Imaging

A dataset containing 91 ultrasound images, corresponding to the oblique axial plane at the level of minimal anteroposterior hiatal (C-plane), from 35 patients was used for validation[5]. All C-planes were selected by the same operator. The dataset had 35 images acquired during Valsalva, 20 images during contraction and 36 images at rest to cover all the stages during a standard diagnosis with some extreme cases and large anatomical variability. Images had a mean pixel size and standard deviation (SD) of 0.54±0.07 mm, with variable image resolutions ([199-286]×[176-223] pixels, for width and length, respectively). All 91 images were manually segmented by 3 different operators with at least 6 months of experience in evaluating pelvic floor 3D ultrasound images. Each operator segmented each image only once. More details on the dataset can be found in the work of Sindhwani et al.[5].

## 3.2 Implementation details

For the purpose of this study, all original US images were automatically cropped or padded to 214x262 pixels primarily for normalisation and removing unnecessary background. In training, for the SU-Net and U-Net, we used a mini-batch size of 32 images, we linearly resized the data to 107x131 pixels and used a data augmentation strategy by applying an affine transformation with 6 degrees-of-freedom. The number of channels was fixed to 64. For the SU-Net with SELU-dropout, a dropout rate of 0.5 was used. During training, the images and labels from the three operators were both shuffled before feeding into respective mini-batches. The networks were implemented in TensorFlow[33] and trained with an Adam optimiser[34] with a learning rate of 0.0001, on a desktop with a 24GB NVIDIA Quadro P6000. For each automatic segmentation obtained, post-processing morphological operators to fill holes (i.e., flood fill of pixels that cannot be reached from the boundary of the image) and remove unconnected regions by selecting the region with the largest area, were also applied. For the U-Net+DC and ResNet we used a mini-batch size of 10, 128 initial channels and a learning rate of 0.001 (all the rest of hyper-parameters, pre-processing and post-processing were kept the same).

## 4 Results

First, using the three manual labels available for each image as a ground truth, we evaluated the performance of the proposed network using the pairwise comparison strategy defined in Section 2.5 with the metrics described in Section 2.4 . For comparison purposes, we also report the results obtained with the baseline U-Net architecture, and the U-Net+DC and ResNet architectures. Median values and standard deviations for each metric are shown in **Table 1**. Statistical analysis comparing the mean values for each image (average of the operators) obtained with the U-Net and the SU-Net showed a statistically significant difference for the Dice, Jaccard, Hausdorff, *SMAD* and *FPD* metrics (*p-values=0.030, 0.022, 0.004, 0.027, 0.031,* respectively), and no significant difference for *MAD* an *FND* metrics (*p-values=0.064, 0.183,* respectively). However, when comparing the values of all metrics using SELU-dropout and without SELU-dropout, no statistically significant difference was found (all *p-values>0.37*). Furthermore, no statistically significant difference was found when comparing the SU-Net and U-Net+DC (all *p-values>*0.30), or when comparing the SU-Net with ResNet (all *p-values>*0.08). Differences between the three operators (i.e., interoperator differences), not considering the automatic segmentations, are reported using the same metrics and shown in **Table 2**. Williams' indices are reported in Table 3 to compare the agreement between automatic and manual segmentations with the agreement among manual segmentations using the metrics described in Section 2.4 .

**Table 1** Performance of the SU-Net, SU-Net+dropout, U-Net, U-Net+DC and ResNet networks by employing a pairwise comparison with the three manual labels available for each ultrasound image. This table also contains results from a previous study (Sindhwani et al.[5]). Results are reported using median [interquartile range].

| Method | Dice | Jaccard | Hausdorff (in mm) | *MAD* (in mm) | *SMAD* (in mm) | *FPD* | *FND* |
|---|---|---|---|---|---|---|---|
| ***SU-Net*** | 0.90 [0.08] | 0.82 [0.12] | 4.21 [3.92] | 1.19 [1.15] | 1.16 [1.02] | 0.07 [0.13] | 0.09 [0.16] |
| SU-Net+dropout | 0.90 [0.08] | 0.81 [0.13] | 3.90 [3.83] | 1.21 [1.16] | 1.23 [1.09] | 0.07 [0.13] | 0.09 [0.16] |
| U-Net | 0.89 [0.11] | 0.80 [0.18] | 4.49 [5.67] | 1.31 [1.42] | 1.34 [1.41] | 0.07 [0.16] | 0.08 [0.16] |
| U-Net+DC | 0.90 [0.08] | 0.82 [0.13] | 3.97 [3.87] | 1.18 [3.86] | 1.17 [1.23] | 0.05 [0.13] | 0.11 [0.15] |
| ResNet | 0.91 [0.08] | 0.83 [0.14] | 3.59 [4.22] | 1.13 [1.14] | 1.10 [1.07] | 0.06 [0.14] | 0.07 [0.13] |
| Sindhwani et al.[5] | 0.92 [0.05] | 0.85 [0.09] | 5.73 [3.90] | 2.10 [1.54] | - | - | - |

**Table 2** Differences between the manual labels from the three operators (i.e., inter-observer differences). Results are reported using median [interquartile range].

| Dice | Jaccard | Hausdorff (in mm) | *MAD* (in mm) | *SMAD* (in mm) | *FPD* | *FND* |
|---|---|---|---|---|---|---|
| 0.92 [0.06] | 0.85 [0.10] | 3.05 [2.33] | 1.01 [0.85] | 1.01 [0.81] | 0.03 [0.08] | 0.08 [0.15] |

**Table 3** Williams' indices (WI) [95% CI] for the SU-Net, SU-Net+dropout, U-Net, U-Net+DC and ResNet architectures for each evaluation metric. A CI containing the value 1.0 indicates a good agreement between the automatic method and the three operators.

| Method | WI Dice | WI Jaccard | WI Hausdorff (in mm) | *WI MAD* (in mm) | *WI SMAD* (in mm) | *WI FPD* | *WI FND* |
|---|---|---|---|---|---|---|---|
| ***SU-Net*** | 1.032 [1.03,1.03] | 1.052 [1.05,1.06] | 0.677 [0.67,0.69] | 0.738 [0.73,0.75] | 0.776 [0.77,0.79] | 0.425 [0.40,0.45] | 0.588 [0.57,0.61] |
| SU-Net+dropout | 1.032 [1.03,1.03] | 1.051 [1.05,1.05] | 0.701 [0.69,0.71] | 0.751 [0.74,0.76] | 0.784 [0.77,0.80] | 0.420 [0.40,0.44] | 0.591 [0.57,0.62] |
| U-Net | 1.085 [1.08,1.09] | 1.111 [1.10,1.12] | 0.530 [0.52,0.54] | 0.577 [0.56,0.59] | 0.538 [0.52,0.56] | 0.281 [0.26,0.30] | 0.439 [0.42,0.46] |
| U-Net+DC | 1.033 [1.03,1.04] | 1.053 [1.05,1.06] | 0.712 [0.70,0.72] | 0.723 [0.71,0.74] | 0.756 [0.74,0.77] | 0.395 [0.37,0.42] | 0.706 [0.69,0.72] |
| ResNet | 1.037 [1.03,1.04] | 1.061 [1.06,1.07] | 0.717 [0.71,0.73] | 0.726 [0.71,0.74] | 0.731 [0.72,0.74] | 0.533 [0.50,0.57] | 0.52 [0.5,0.54] |

**Table 4** shows the mean differences in area of the segmented regions in terms of computer-to-operator differences and interoperator differences during the three different stages, and with the corresponding Williams' indices testing the performances.

**Table 4** Computer-to-operator differences in the segmented area when comparing the automatic and manual segmentations (COD) with SU-Net and when comparing the manual segmentations between operators (IOD), with the corresponding Williams' indices and the 95% CI. Results are reported using mean (±SD).

| Stage | Contraction | Valsalva | Rest |
|---|---|---|---|
| COD | 0.62±0.91 | 0.86±1.89 | 0.60±1.22 |
| IOD | 0.52±0.70 | 0.62±1.03 | 0.61±0.92 |
| Williams' Index | 0.80 | 0.72 | 0.85 |
| [95% CI] | [0.72,0.89] | [0.68,0.76] | [0.80,0.90] |

**Fig. 3** shows examples of original images with the corresponding segmentation results obtained with the automatic method together with the three manual labels used as a ground truth, and **Fig. 4** shows examples at the three different stages: rest, Valsalva and during contraction.

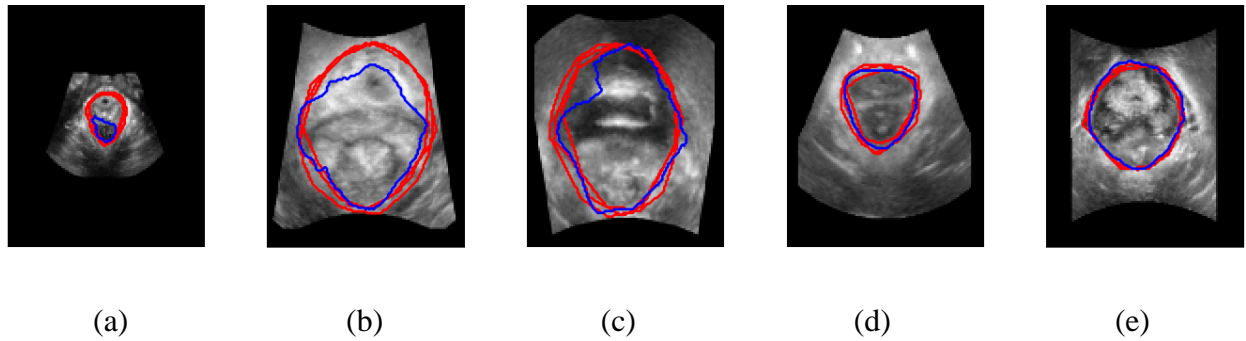

(a)     (b)     (c)     (d)     (e)

**Fig. 3** Segmentation of the levator hiatus using with the SU-Net architecture (blue) compared with the three manual labels (red) for the following percentiles of the Dice coefficient: $0^{th}$ (a), $25^{th}$ (b), $50^{th}$ (c), $75^{th}$ (d) and $100^{th}$ (e).

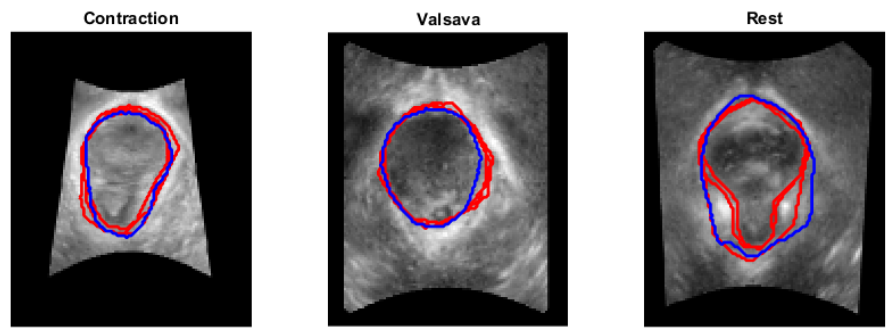

**Fig. 4** Segmentation examples of the levator hiatus at the three different stages (contraction, Valsalva and rest) using the proposed method (blue) compared to the outlines provided by the operators (red). Cases were chosen at the $75^{th}$ percentile of the mean Dice coefficient considering the three operators.

**Fig. 5** shows the histogram of the values obtained after the last SELU at different iterations. **Fig. 6** shows how the dice coefficient converges using the U-Net and SU-Net architectures, and **Fig. 7** the learning curves of the training loss for the U-Net and SU-Net methods.

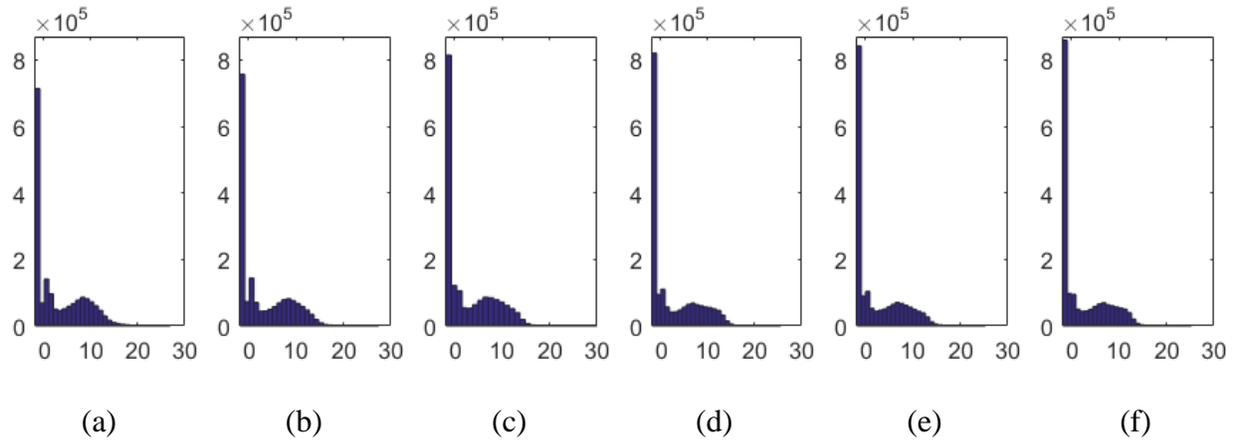

(a) (b) (c) (d) (e) (f)

**Fig. 5** Histogram of the SELU activations at the last block after 500 (a), 1000 (b), 1500 (c), 2000 (d), 2500 (e) and 3000 (f) iterations.

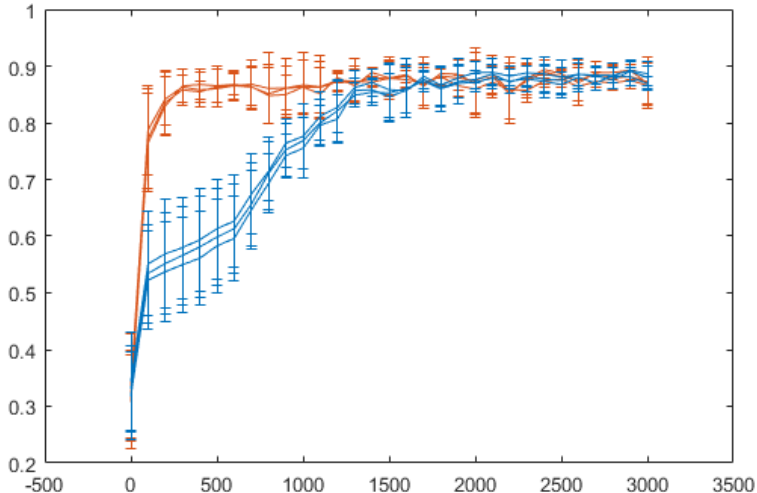

**Fig. 6** Overlap at different iterations (0-3000) for the U-Net (blue) and SU-Net (orange) architectures during testing for the first fold and for the three operators.

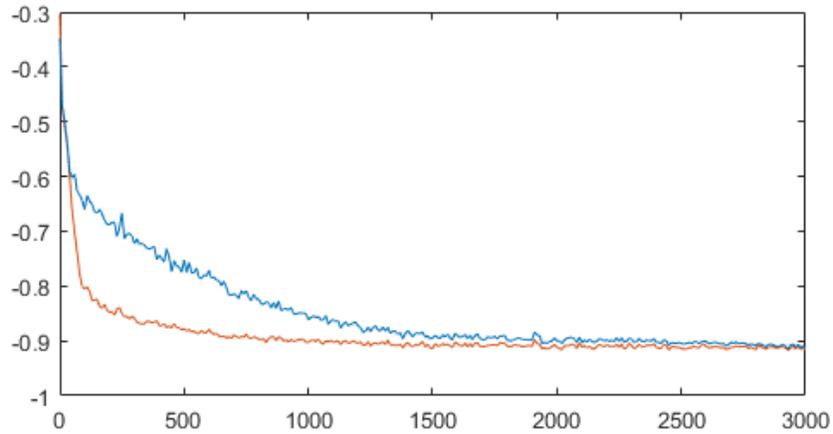

**Fig. 7** Learning curves of the training loss for the U-Net (blue) and SU-Net (orange) architectures averaged for all folds at different iterations (0-3000).

## 5  Discussion

The task of segmenting ultrasound images can be challenging and often results in high variability between operators. In this work, we have presented a fully automatic method, using a convolutional neural network, to segment the pelvic floor levator hiatus on a 2D image plane extracted from a 3D ultrasound volume. A large number of female patients may potentially benefit globally from this approach. We have adopted a recently proposed self-normalising neural network, which for the first time, has been applied in medical imaging to tackle a clinically important application, obtaining either superior or equivalent segmentation results compared to a number of state-of the-art network architectures with clear additional benefits in terms of complexity and memory requirements. Furthermore, based on a set of rigorous statistical tests with real clinical image data, the proposed fully automatic method achieved an equivalent accurate segmentation result compared to the only previous (semi-automated) study presented by Sindhwani et al[5].

State-of-the-art deep-learning architectures have been shown to perform well in the task of segmentation. To the best of our knowledge, this is the first work in medical imaging to replace the batch normalisation with a SELU unit. SNN networks are able to retain many layers with stable training, particularly with a strong regularisation that is advantageous for ultrasound image segmentation. Furthermore, using SELU has the opportunity of reducing the GPU memory requirement and relax the dependency of mini-batch.

We show that the method presented outperformed the U-Net-based architecture by considering region-based and contour-based metrics and confirmed by statistical tests. Although the effective difference, i.e. effect size, is relatively small and subject to further investigation in determining the clinical relevance, SELU may have provided a faster convergence (**Fig. 6** and **Fig. 7**). Furthermore, although it is difficult to draw quantitative conclusion on the efficacy of the SELU units, the activation output distributions shown in **Fig. 5** illustrate the desirably stable variation during training[19]. On the other hand, no statistical significant difference was found when SELU-dropout, U-Net+DC or ResNet were used. Therefore, SELU can potentially provide equivalent or improved results without the mini-batch size limitation.

Comparing the computer-to-observer differences (**Table 1**) with interoperator differences (**Table 2**), we show highly similar results on the median values, however, Williams' indices confidence intervals show that the automatic method strongly agrees with the observers in terms of Dice and Jaccard coefficient with a value very close to 1, but it is not the case for the distance metrics. This result may be due to a disagreement on local parts of the boundaries as shown in **Fig. 3 (c)**, which gives a higher Hausdorff distance value, or due to a larger part of the boundary in disagreement with the operators as shown in **Fig. 3 (b)** which results in a higher *SMAD* value.

As a clinically relevant metric, we evaluated the differences in area at three different stages (contraction, Valsalva and rest). In this case, Williams' indices were smaller than 1, showing some level of disagreement with the operators (**Table 4**). We believe that the results can be further improved by increasing the number of images during training, as the current dataset size is limited and contains some extreme cases with a high variability.

Compared to a previous study[5], in which at least three anatomical points have to be manually identified on the C-plane, we proposed a fully-automatic segmentation algorithm that is able to segment the pelvic floor on the C-plane without operator input of any form, achieving comparable accuracy. Note that, the previous study already achieved competitive results obtaining a good agreement with the three operators (**Table 1** and **Table 2**) and demonstrated to be clinically useful. Furthermore, compared to a solution that requires human interaction (i.e. manual definition of several anatomical landmarks), fully-automatic methods, such as the one proposed in this work, have significant advantages including minimising subjective factors due to intra- and inter observer variations, simpler clinical workflow with minimal uncertainty and quantifiable, repeatable procedure outcome.

The limitation of this work, from a clinical application perspective, is the need to identify the C-plane from a 3D ultrasound volume, which is currently done manually. We have focused on the task of automatically segmenting the pelvic floor on the C-plane mainly for three reasons: 1) the levator hiatus is a mostly flat structure and there is no envisaged clinical benefit of performing a 3D segmentation rather than a 2D one in the C-plane; 2) validation of 2D segmentation results in the same volume but on different C-planes is problematic as it requires to compare manual contours on potentially different images; and 3) the proposed method is meant to be one step of a minimally interactive workflow for pelvic floor disorder analysis. The current work aims at demonstrating the performance of the proposed automatic method in a controlled problem domain, (i.e. where the C-plane is provided), before pursuing more end-to-end solutions. After the successful development reported in this work, we plan to investigate the feasibility of implementing the complete analysis pipeline in which a) the identification of the C-plane would be automated but potentially refined by the user; b) the proposed automated deep-learning based segmentation could be possibly manually refined using an approach similar to that of Wang et al.[12,13] but requiring less user-time than that of Sindhwani et al.[5]; and c) an automated prediction of clinically-relevant measurements and decision support information would be performed based on the user-validated C-plane and levator hiatus.

# 6 Conclusion

In this work, we present a deep learning method based on a self-normalising neural network to automate the process of segmenting the pelvic floor levator hiatus in a 2D plane extracted from an ultrasound volume, which outperforms the equivalent U-Net architecture and foregoes the need for batch normalisation. Compared to previous work, this method is fully automatic with equivalent operator performance in terms of Dice metrics.

*Disclosures*

The authors have no conflict of interest to declare.

*Acknowledgments*

The authors would like to thank Dr. Friyan Tuyrel, Dr. Ixora Atan for providing the data and the manual ground truth labels used in this study. This work was supported by the Wellcome / EPSRC [203145Z/16/Z, WT101957, NS/A000027/1]; and the Royal Society [RG160569].

**Ester Bonmati** is a Research Associate at University College London. She received her BSc, MSc and PhD degrees from University of Girona. Her current research interests include image guided interventions, medical image processing and surgical planning and navigation.

**Yipeng Hu** is a Senior Research Associate at University College London. He received his BEng from Sichuan University and his MSc and PhD from University College London. His current research interests include image-guided interventions and medical image computing.

**Nikhil Sindhwani** is a Technical Product manager at Nobel Biocare, Belgium. He received his B.Tech from SASTRA University, his MS from University of Illinois and his PhD from KU Leuven.

**Hans Peter Dietz** is a Professor and an Obstetrician and Gynaecologist at University of Sydney. He has more than 25 years of expertise in pelvic floor assessment and pelvic organ prolapse, being a pioneer in the field. His research interests include the effect of childbirth on the pelvic floor, diagnostic ultrasound imaging techniques in urogynaecology and the prevention and treatment of pelvic floor disorders.

**Jan D'hooge** is a Professor at the department of cardiology at KU Leuven. He received his MS and his PhD from the KU Leuven. His current research interests include myocardial tissue characterisation and strain imaging in combination with cardiac patho-physiology.

**Dean Barratt** holds the position of Reader in Medical Image Computing at University College London. He received his undergraduate degree from Oxford University and his MSc and PhD from the Imperial College London. His research interests include technologies to enable image-guided diagnosis and therapy, with a particular interest in multimodal image registration, image segmentation, computational organ motion modelling, interventional ultrasound, and cancer.

**Jan Deprest** is a Professor at KU Leuven, head of the Centre for Surgical Technologies, chair of the department of Development and Regeneration and head of Organ Systems. His current research interests include fetal and perinatal medicine and image-guided intrauterine minimally invasive fetal diagnosis and therapy.

**Tom Vercauteren** is a Senior Lecturer in Interventional Imaging at University College London and the Deputy Director for the Wellcome / EPSRC Centre for Interventional and Surgical Sciences (WEISS). He graduated from Ecole Polytechnique, received his MSc from Columbia

University and his PhD from Inria and Ecole des Mines de Paris. His current research interests include medical image computing and interventional imaging devices.

**Caption List**

Fig. 1 Network architecture, where $S_1$ and $S_2$ correspond to the spatial dimension and *nc* to the number of channels. For the U-Net, the SELU unit is replaced by batch normalisation and ReLU, and for the U-Net with dilated convolution (U-Net+DC), the last layer is also replaced by a dilated convolution.

Fig. 2 SU-Net architecture (a) versus a U-Net architecture (b).

Fig. 3 Segmentation of the levator hiatus using with the SU-Net architecture (blue) compared with the three manual labels (red) for the following percentiles of the Dice coefficient: $0^{th}$ (a), $25^{th}$ (b), $50^{th}$ (c), $75^{th}$ (d) and $100^{th}$ (e).

Fig. 4 Segmentation examples of the levator hiatus at the three different stages (contraction, Valsalva and rest) using the proposed method (blue) compared to the outlines provided by the operators (red). Cases were chosen at the $75^{th}$ percentile of the mean Dice coefficient considering the three operators.

**Fig. 5** Histogram of the SELU activations at the last block after 500 (a), 1000 (b), 1500 (c), 2000 (d), 2500 (e) and 3000 (f) iterations.

**Fig. 6** Overlap at different iterations (0-3000) for the U-Net (blue) and SU-Net (orange) architectures during testing for the first fold and for the three operators.

**Fig. 7** Learning curves of the training loss for the U-Net (blue) and SU-Net (orange) architectures averaged for all folds at different iterations (0-3000).

**Table 1** Performance of the SU-Net, SU-Net+dropout, U-Net, U-Net+DC and ResNet networks by employing a pairwise comparison with the three manual labels available for each ultrasound image. This table also contains results from a previous study (Sindhwani et al.[5]). Results are reported using median [interquartile range].

Table 2 Differences between the manual labels from the three operators (i.e., inter-observer differences). Results are reported using median [interquartile range].

Table 3 Williams' indices (WI) [95% CI] for the SU-Net, SU-Net+dropout, U-Net, U-Net+DC and ResNet architectures for each evaluation metric. A CI containing the value 1.0 indicates a good agreement between the automatic method and the three operators.

Table 4 Computer-to-operator differences in the segmented area when comparing the automatic and manual segmentations (COD) with SU-Net and when comparing the manual segmentations between operators (IOD), with the corresponding Williams' indices and the 95% CI. Results are reported using mean (±SD).